\documentclass{article}

\usepackage{PRIMEarxiv}

\usepackage[utf8]{inputenc} 
\usepackage[T1]{fontenc}    
\usepackage{hyperref}       
\usepackage{url}            
\usepackage{booktabs}       
\usepackage{amsfonts}       
\usepackage{nicefrac}       
\usepackage{microtype}      
\usepackage{lipsum}
\usepackage{fancyhdr}       
\usepackage{graphicx}       
\graphicspath{{media/}}     
\usepackage{amsmath}
\usepackage{subcaption}

\title{Evaluation of Vision Transformers for Multimodal Image Classification: A Case Study on Brain, Lung, and Kidney Tumors}

\author{
  Óscar A. Martín \\
  Centro de Tecnologías de la Imagen (CTIM)\\
  Instituto Universitario de Cibernética, Empresas y Sociedad (IUCES)\\
  University of Las Palmas de Gran Canaria \\
  35017 Las Palmas de Gran Canaria, Spain\\
  \texttt{oscar.martin104@alu.ulpgc.es} \\
   \And
  Javier Sánchez \\
  Centro de Tecnologías de la Imagen (CTIM)\\
  Instituto Universitario de Cibernética, Empresas y Sociedad (IUCES)\\
  University of Las Palmas de Gran Canaria \\
  35017 Las Palmas de Gran Canaria, Spain\\
  \texttt{jsanchez@ulpgc.es} \\
}

\begin{document}
\maketitle

\begin{abstract}
Neural networks have become the standard technique for medical diagnostics, especially in cancer detection and classification. This work evaluates the performance of Vision Transformers architectures, including Swin Transformer and MaxViT, in several datasets of magnetic resonance imaging (MRI) and computed tomography (CT) scans. We used three training sets of images with brain, lung, and kidney tumors. Each dataset includes different classification labels, from brain gliomas and meningiomas to benign and malignant lung conditions and kidney anomalies such as cysts and cancers. This work aims to analyze the behavior of the neural networks in each dataset and the benefits of combining different image modalities and tumor classes. We designed several experiments by fine-tuning the models on combined and individual datasets. The results revealed that the Swin Transformer provided high accuracy, achieving up to 99\% on average for individual datasets and 99.4\% accuracy for the combined dataset. This research highlights the adaptability of Transformer-based models to various image modalities and features. However, challenges persist, including limited annotated data and interpretability issues. Future work will expand this study by incorporating other image modalities and enhancing diagnostic capabilities. Integrating these models across diverse datasets could mark a significant advance in precision medicine, paving the way for more efficient and comprehensive healthcare solutions.
\end{abstract}


\section{Introduction}
Cancer is a term that encompasses a broad group of diseases. The statistics show that it is the leading cause of death affecting society, resulting in millions of deaths each year. According to~\cite{sung2021global}, there were an estimated 19.3 million new cases of cancer and 9.9 million cancer-related deaths in 2020.

In the last decade, medical technology has rapidly improved, with neural networks playing a key role. The intersection of computer science and medicine has yielded significant improvements in disease research and diagnosis. Artificial intelligence (AI) has lightened the burden on medical professionals, enhancing diagnostic efficiency. AI finds applications in diverse areas, from health-monitoring wearables to DNA analysis algorithms.

Technology plays a crucial role in early and accurate cancer diagnoses, providing healthcare professionals with automated tools to detect diseases efficiently and minimizing human error and workload. In particular, the use of deep neural networks for the detection and classification has become widely used in several types of tumors, such as brain~\cite{deepak2019brain}, lung~\cite{coudray2018classification}, or kidney tumors~\cite{hermsen2019deep}. Nevertheless, not many studies have assessed the performance of these neural networks in detecting and classifying different types of diseases using a single network and the benefits this can bring.  

This work specifically investigates the use of recent models for image-based diagnosis, particularly using magnetic resonance imaging (MRI) and computed tomography (CT) scans to detect tumors. We explore architectures based on the Vision Transformer and its variants, which excel at processing images and identifying anomalous patterns. Our work aims to classify different types of tumors with distinct image formats into a single model. This will allow us to assess the capabilities of these architectures in generalizing to multiple diseases and image formats. The objective is to integrate different data and image formats into a single Transformer model, adapting to cancerous and benign disease detection. The Transformer architecture has been successfully applied in computer vision, superseding convolutional neural networks (CNNs) in many tasks. The differences between CNNs and Transformers are notable from an architectural perspective, and the choice between them will depend on the specific task. We analyze several architectures, such as the Vision Transformer (ViT) ~\cite{vit}, the Swin Transformer~\cite{swin,swin2}, and MaxVit~\cite{maxvit}. 

For the training of these networks, we selected three datasets: one of brain images~\cite{brain-dataset}, containing three types of tumors (\textit{gliomas}, \textit{meningiomas}, and \textit{pituitary tumors}); another of lung images~\cite{lung-dataset-article,lung-dataset}, containing CT scans of lung cancer and healthy lungs, with three distinct classes (\textit{normal}, \textit{benign}, and \textit{malignant}); and another dataset of kidney images~\cite{kidney-dataset}, containing four different categories (\textit{non-tumor}, \textit{stone}, \textit{cyst}, and \textit{cancer}). The first dataset includes 15,000 MRIs, with 5,000 images in each subclass; the second dataset contains 1,190 CT scans of lung cancer and healthy lungs; and the third one has 12,446 CT scans distributed in four classes. 

The experimental results analyze the performance of each neural network when trained from scratch and using transfer learning. In many cases, the models' accuracy is above 99\%, with the Swin Transformer providing outstanding results, followed by MaxViT. These promising results demonstrate that vision Transformers can play a key role in medical image analysis.

The main contribution of this work lies in its systematic evaluation of multiple ViT variants across heterogeneous medical imaging datasets and different tumor types and modalities. Unlike prior studies that typically focus on a single organ or imaging modality, our approach explores the potential of ViTs to generalize within a unified framework, highlighting their adaptability in multiclass, multimodal classification tasks. By comparing model performance on individual datasets versus a combined dataset, we offer novel insights into how the models leverage shared patterns and modality-specific features through their self-attention mechanisms. 

Section~\ref{se:relate_work} summarizes the state-of-the-art works on brain, lung, and kidney tumor detection methods. Section~\ref{se:methods} details the datasets used in this work and the neural networks employed in the classification task. The results in Section~\ref{se:results} assess the performance of the neural networks for each dataset and the accuracy by combining the three datasets in a single set. Finally, the conclusions in Section~\ref{se:conclusion} summarize the main ideas and contributions of the work and propose some ideas for future works. 

\section{Related Work}
\label{se:relate_work}

Artificial intelligence is making significant advances in tumor detection, with deep learning models, like EfficientNet~\cite{medina2023high}, achieving high accuracy rates with MRI images. Other standard models, such as VGG and MobileNet, have also shown outstanding results~\cite{reyes2024performance}. Furthermore, the use of Transformer models, such as the Vision and the Swin Transformers, are being explored to overcome the limitations of CNNs in medical image classification and segmentation. These models can capture global relationships in images, which is crucial for accuracy in medical diagnosis.

A recent work~\cite{khan2023recentsurveyvisiontransformers} highlights the importance of segmenting medical images based on Vision Transformers instead of convolutional networks. The latter are effective in capturing local correlations, although they are limited in capturing global relationships. 

\subsection{Brain tumors}
Brain tumor classification has received significant attention in the last few years, particularly through the analysis of MR images. Various studies have been conducted to enhance the performance of brain tumor classification using different methodologies. The work presented in ~\cite{cheng2015enhanced} focused on brain tumor classification based on T1-weighted contrast-enhanced MRI. They proposed a dataset that has been used in many subsequent works.  

Traditional techniques typically classify images based on two main steps: feature extraction and classification. The features proposed in~\cite{gumaei2019hybrid} were based on PCA and GIST techniques, and the classification was carried out through a regularized extreme learning machine. The last step can also involve the use of neural networks, like the work proposed in \cite{ismael2018brain}, which relies on the 2D Discrete Wavelet Transform (DWT) and 2D Gabor filters to extract statistical features from MRI and then feed them into a neural network. The approach presented in~\cite{bahadure2018comparative} tackled the problem of segmentation and classification of MRI using genetic algorithms. The authors of~\cite{afshar2018brain} explored brain tumor classification using Capsule Networks. This type of network has several benefits over CNNs since they are robust to rotation and affine transformations and require less training data.

The system proposed in~\cite{sajjad2019multi} consists of three main steps: tumor segmentation, data augmentation, and deep feature extraction and classification. It relied on extensive data augmentation techniques and the fine-tuning of a VGG-19 network. Many works \cite{ayadi2021deep} have extensively used pure CNN models for brain tumor classification, obtaining high accuracy in different datasets. 

Several works~\cite{srinivas2022deep} have conducted a performance analysis of transfer learning in CNN models (such as VGG-16, ResNet-50, or Inception-v3) for automatic prediction of tumor cells in the brain. A recent work~\cite{reyes2024performance} reported a detailed performance assessment analysis of many convolutional architectures from different perspectives, drawing important conclusions about these techniques. Several models can attain high accuracy, even for networks with a relatively low number of parameters like EfficientNet~\cite{medina2023high}. 

\subsection{Lung tumors}
    
Lung cancer remains one of the leading causes of cancer-related deaths worldwide. Early and accurate detection through imaging techniques like CT scans is crucial for improving patient outcomes. Deep learning models have shown significant promise in automating the classification of lung tumors from CT scans. 

Classification of lung tumors using machine learning techniques has been an active area of research. Numerous studies have explored algorithms and methodologies to improve diagnostic accuracy and efficiency. 

Early work in lung tumor classification primarily utilized traditional machine learning algorithms such as Support Vector Machines (SVM), k-Nearest Neighbors (k-NN), and Decision Trees. For instance, the authors of~\cite{7158331} employed SVM for classifying lung nodules in CT images, achieving notable accuracy by optimizing hyperparameters and using feature selection techniques to reduce dimensionality. Similarly, \cite{sathishkumar2019detection} used k-NN combined with SVM, demonstrating improved classification performance on small datasets.

With the advent of deep learning, CNNs have become the preferred choice for image-based lung tumor classification. The work presented in \cite{shen15} developed a deep CNN model that outperformed traditional methods by automatically learning hierarchical features from raw CT images. Their approach significantly reduced the need for manual feature extraction, leading to higher accuracy and robustness. On the other hand, \cite{LIU2018262} further advanced this field by introducing a multi-view CNN that integrates information from multiple CT slices, enhancing the ability of the model to capture spatial dependencies and improve classification accuracy. This method demonstrated superior performance in distinguishing between benign and malignant nodules compared to single-view CNNs.

Ensemble methods, which combine the predictions of multiple models, have been explored to enhance classification performance. For example, an ensemble of CNNs and several traditional machine-learning techniques to classify lung tumors~\cite{zhang2019ensemble}, achieving improved accuracy and robustness by mitigating the weaknesses of individual models.

Transfer learning, which involves fine-tuning pre-trained models on specific datasets, has gained popularity due to its effectiveness in scenarios with limited labeled data. The model presented in \cite{8624570} employed transfer learning using pre-trained CNNs, achieving high classification accuracy with reduced training time. Their approach demonstrated that leveraging pre-trained models can enhance performance, especially when dealing with small medical datasets.

Recent studies have focused on improving the interpretability and explainability of machine learning models in lung tumor classification. A recent work~\cite{WANG2024105646} introduced an explainable AI framework that combines a 3D U-Net with attention mechanisms to highlight the most relevant regions in CT images.

\subsection{Kidney tumors}

The application of machine learning to the classification of kidney tumors has seen significant advancements in recent years. Various methods and technologies have been developed to aid in the early detection and accurate segmentation of kidney tumors. For instance, the work in~\cite{kim04} utilized a computer-aided detection system for kidney tumors on abdominal CT scans, employing a gray-level threshold method for segmentation and texture analysis for tumor detection. Similarly, \cite{zhou2016atlas} presented a semi-automatic kidney tumor detection and segmentation method using atlas-based segmentation.

The classification of kidney tumors into subtypes such as benign, malignant, and histological categories (e.g., clear cell renal carcinoma, papillary renal cell carcinoma) is crucial for guiding treatment strategies. Early approaches often utilized traditional techniques, such as SVMs and Random Forests, relying on manually extracted features such as texture, shape, and intensity descriptors. For example, \cite{feng2018machine} explored handcrafted features combined with SVMs for detecting renal masses, achieving promising results in sensitivity and specificity. Similarly, \cite{erdim2020prediction} utilized radiomics features in combination with eight machine-learning techniques, like logistic regression, decision trees, and SVMs, to classify renal tumors.

Advancements in deep learning techniques have also played a crucial role in kidney tumor detection. For example, \cite{rathnayaka2019kidney} proposed a CNN-based U-Net architecture with an attention mechanism for kidney tumor detection. On the other hand, \cite{lin2021automated} developed a 3D U-Net-based deep convolutional neural network for automatically segmenting kidney and renal masses, achieving high accuracy in tumor segmentation. Furthermore, \cite{alzu2022kidney} and \cite{praveen2023resnet} focused on deep learning approaches for kidney tumor detection and classification, utilizing models such as 2D-CNN, ResNet, and ResNeXt to improve the detection accuracy. The work in~\cite{ozbay2024kidney} explored self-supervised learning for kidney tumor classification on CT images. In addition, \cite{bogomolov2017development} studied the development of an LED-based near-infrared sensor for human kidney tumor diagnostics, providing a simplified alternative to conventional NIR spectroscopic methods.

\section{Materials and Methods}
\label{se:methods}

This section details the datasets and methods employed in our study. Our work is based on two types of medical imaging: computed tomography, which uses X-rays to obtain detailed images, useful in oncology and other medical areas; and magnetic resonance imaging, which produces detailed three-dimensional images using magnetic fields and electromagnetic waves, differentiating tissues, although it is more expensive than tomography.

\subsection{Datasets}
We selected three datasets related to brain, lung, and kidney tumors. In the next sections, we explain each dataset in detail. 

\subsubsection{Brain Tumor Dataset}
Brain tumors originate from the abnormal growth of cancerous cells within the brain. They can be benign or malignant and may affect various areas, leading to symptoms depending on their location, size, and severity. Research indicates that patient survival largely depends on the complete tumor removal, making early detection crucial for timely and appropriate treatment before the tumor grows.

In this work, we will test the classification of the following brain tumors:
\begin{itemize}
    \item \textit{Glioma}: This common tumor type has several variants. Some are considered non-dangerous, but others are quite aggressive, spreading to healthy brain tissues and causing pressure on the brain or spinal cord. Gliomas are cancerous cells resembling glial cells surrounding brain and spinal cord nerve cells. They grow within the white matter and dangerously spread through healthy brain tissues.
    \item \textit{Meningioma}: This is the most common primary brain tumor, classified into three grades based on their characteristics, with the high grade being rare.
        \begin{itemize}
            \item Grade I: The most common, is a low-grade tumor with slow growth.
            \item Grade II: An intermediate grade, known as atypical meningioma, characterized by a higher likelihood of recurrence after removal.
            \item Grade III: The highest grade, less common, characterized as malignant and called anaplastic meningioma. It differs in appearance from normal cells and expands rapidly. Atypical and anaplastic meningiomas can spread throughout the brain and other body organs. A mass on the outer tissue layer of the brain can identify these meningiomas. With this data, the importance of including this cancer type in the dataset for early detection training is evident.
        \end{itemize}
    \item \textit{Pituitary Tumor}: Generally, this type of brain tumor is non-invasive. It is likely to be benign with low growth potential. However, there is a risk of growth and spread to other parts, such as the optic nerve or carotid arteries, presenting more aggressive characteristics. These often go unnoticed and are detected during tests for reasons unrelated to cancer.
\end{itemize}

\begin{figure}[!ht]
    \centering
    \begin{tabular}{ccc}
        \includegraphics[width=5cm,height=5cm,angle=0]{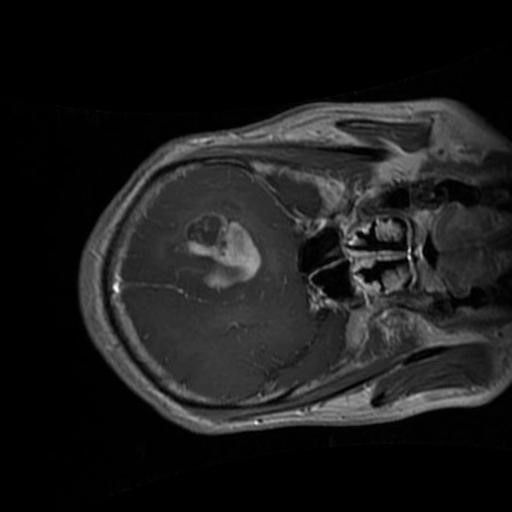} &
        \includegraphics[width=5cm,height=5cm,angle=0]{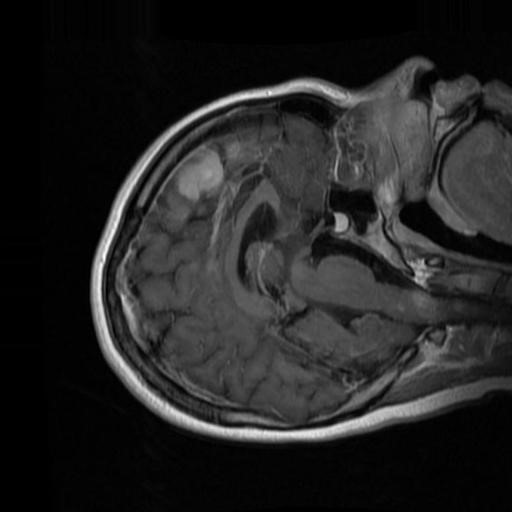} &
        \includegraphics[width=5cm,height=5cm,angle=0]{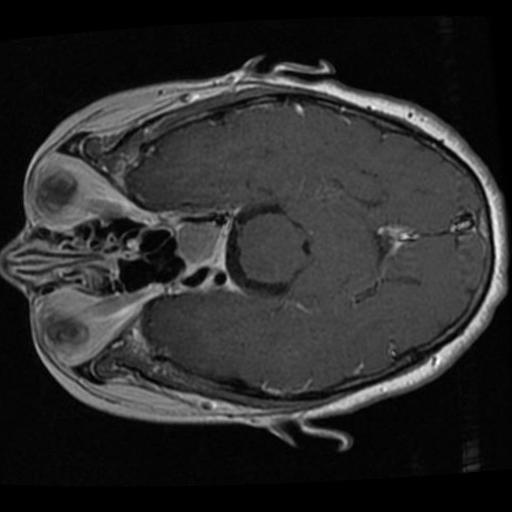} \\
        \includegraphics[width=5cm,height=5cm,angle=0]{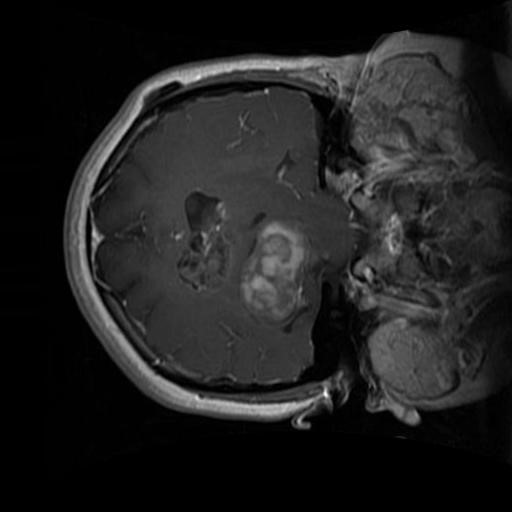} &        \includegraphics[width=5cm,height=5cm,angle=0]{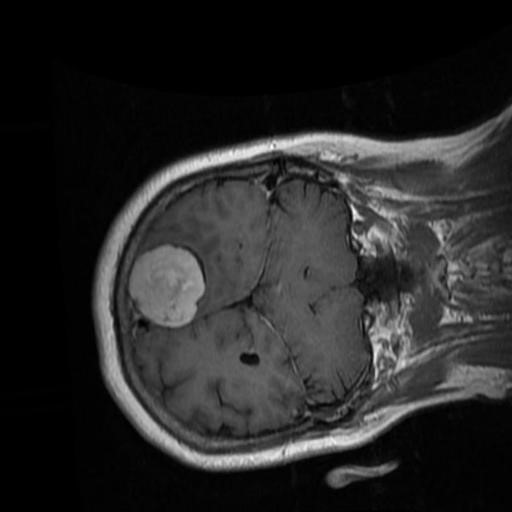} &  
        \includegraphics[width=5cm,height=5cm,angle=0]{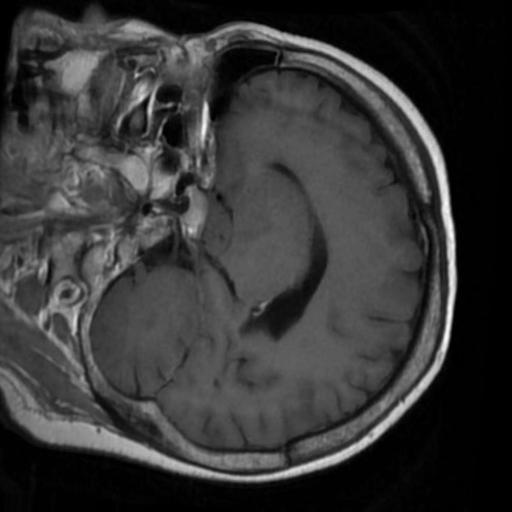} \\ 
    \end{tabular}
    \caption{Samples from the brain dataset: the first column shows two examples of Glioma tumors; the middle column shows Meningioma tumors; and the last column depicts images of Pituitary tumors.}
    \label{fig:brain_tumors}
\end{figure}

The brain tumor dataset used in this study is part of the Multi-Cancer Imaging Dataset~\cite{brain-dataset}. This dataset includes MRI images categorized into the three tumor types. The brain cancer images were obtained from other sources~\cite{cheng2015enhanced,Cheng2017}. 

Figure~\ref{fig:brain_tumors} depicts several examples of brain tumors in the dataset. This dataset forms part of a larger collection that encompasses eight distinct cancer types from different organs: acute lymphoblastic leukemia, brain, breast, cervical, kidney, lung, colon, and oral cancer. The images are in JPEG format with dimensions of $512 \times 512$ pixels. We selected brain cancer, which contains the three previous tumor subclasses. Each subclass includes the same number of images (5,000 MRI).

\subsubsection{Lung Tumor Dataset}
Lung tumors can arise from various lung cells, including the bronchi, bronchioles, and pulmonary alveoli. Lung cancer is a harmful disease that poses a lethal threat to patients’ health. This causes more deaths than breast, colon, and prostate cancers combined. However, the prognosis depends on the type of cancer, its spread, and size. Lung tumors often do not cause symptoms and are usually detected during imaging tests conducted for reasons unrelated to lung cancer. 

Since these frequently go unnoticed in clinical histories, developing a predictive model that can provide information on identifying a potential lung tumor could be beneficial. An early diagnosis of lung cancer, when the tumor has not spread to other tissues, can lead to a favorable prognosis in up to 90\% of patients.

\begin{figure}[!ht]
    \centering
    \begin{tabular}{ccc}
        \includegraphics[width=0.3\textwidth]{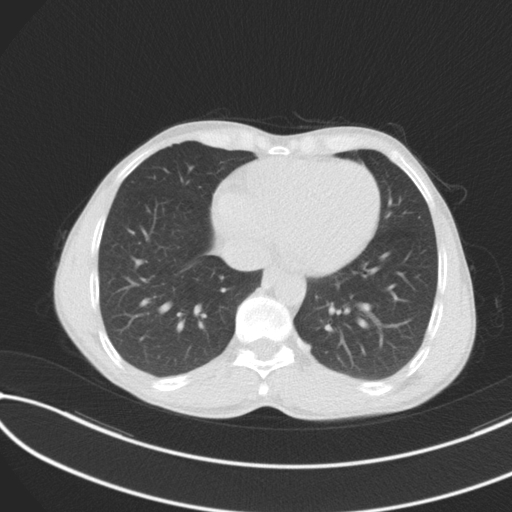} &
        \includegraphics[width=0.3\textwidth]{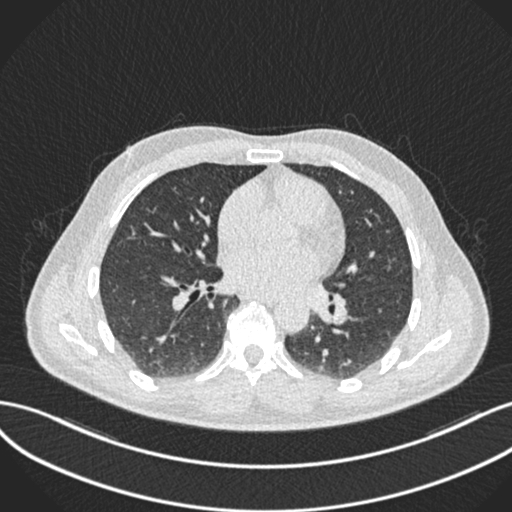} &
        \includegraphics[width=0.3\textwidth]{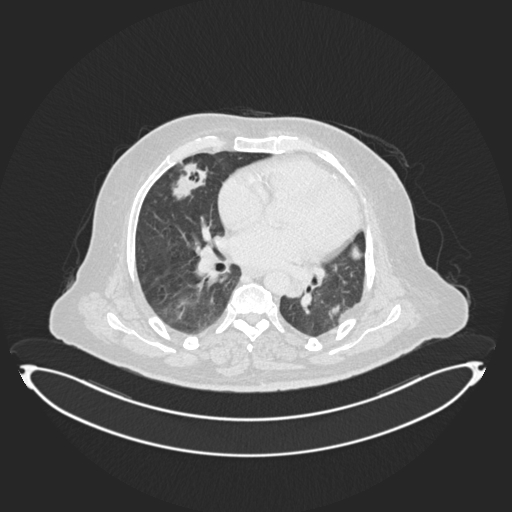}  \\
        \includegraphics[width=0.3\textwidth]{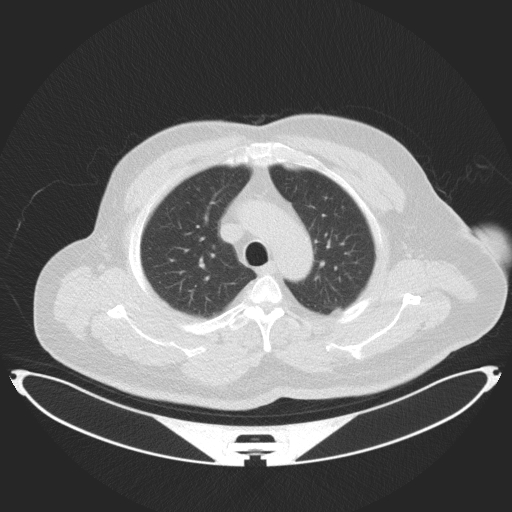} &
        \includegraphics[width=0.3\textwidth]{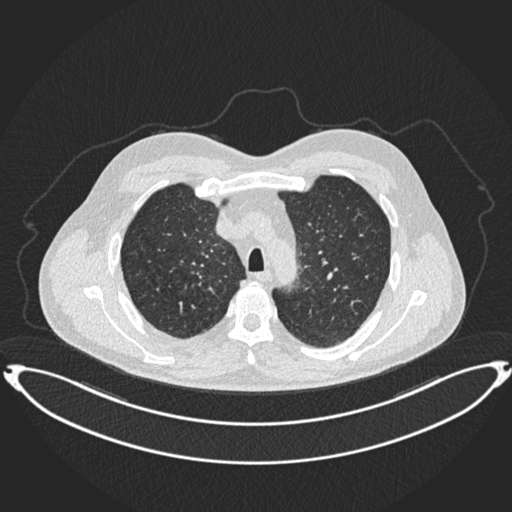} &
        \includegraphics[width=0.3\textwidth]{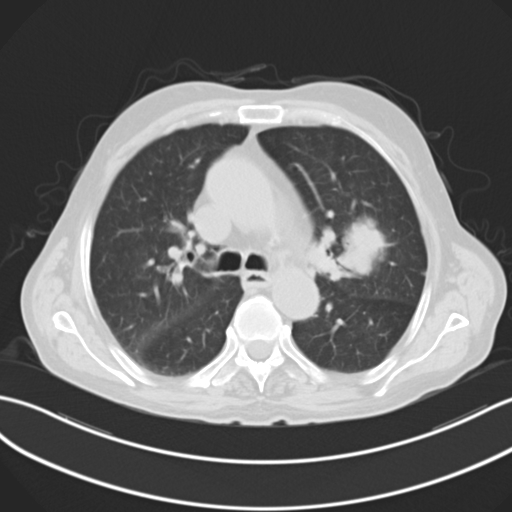}    
    \end{tabular}
    \caption{Examples of lung images. The images in the first column represent lungs with no tumors. The images in the middle column represent images of benign tumors, and the last column contains two images with malignant tumors.}
    \label{fig:lung_tumors}
\end{figure}

We selected a dataset from Kaggle~\cite{lung-dataset-article,lung-dataset}, which includes CT scans of lung cancer and healthy lungs. The dataset has three classes: \textit{normal}, \textit{benign}, and \textit{malignant}. Figure~\ref{fig:lung_tumors} depicts two examples of each class in the dataset. 

The images were in DICOM format and were preprocessed to anonymize patient identifiers. The dataset comprises 110 cases, varying in patient age, sex, residential area, and socioeconomic status, among other variables. It contained 1,190 images, although we detected and removed duplicate files, with a total of 1,054 unique images. These are distributed as shown in Table~\ref{tab:distribution_lung_dataset}.

\begin{table}[!ht]
    \caption{Distribution of samples in the lung dataset. For each type of class, this table shows the number of cases in the second column and the number of images in the third column.}
    \centering
    \begin{tabular}{lcc}
        \toprule
         Type  & Images \\
         \midrule
         Healthy Lung  & 405\\
         Benign Tumor Lung  & 102\\
         Cancerous Tumor Lung  & 547\\
         \bottomrule
    \end{tabular}
    \label{tab:distribution_lung_dataset}
\end{table}

\subsubsection{Kidney Dataset}
In the United States, approximately 430,000 individuals were diagnosed with kidney cancer in the year 2020, and the number of cases has been increasing over the decades. It ranks as the sixth most common type of cancer among men and ninth among women. The relative survival rate is 77\% after five years, depending on various factors such as treatment, cancer spread, and patient age. This statistic refers to the patient's life expectancy during the stipulated time after the diagnosis of the disease or the start of treatment.

We selected a dataset of kidney images from Kaggle~\cite{kidney-dataset}, classified into four different categories:

\begin{itemize}
    \item \textit{Non-tumor}: This category represents images of healthy kidneys.
    \item \textit{Stone}: This category represents images where an abnormal solid material is located in the kidney. This type of material occurs in the kidneys when levels of certain minerals are high. It usually presents with sharp pain in the lower back or groin or even blood in the urine. Diagnosis depends on physical health, laboratory tests, and the size of the stone visible in the images. Generally, treatment involves removing or breaking the stone into pieces.
    \item \textit{Cyst}: This class represents images where small aqueous or liquid structures are identified and are not normally harmful to health. It is necessary to distinguish between a simple cyst, which has no risk of turning into kidney cancer, and a complex cyst, which does have a risk of becoming kidney cancer, even though this risk is low and would require careful diagnosis by a urological surgeon.
    \item \textit{Cancer}: This represents images with a cancerous mass. Renal cell carcinoma is the most common type of tumor, accounting for up to 90\% of kidney cancer. Many cases are detected in the early stages of the tumor. Generally, it does not present symptoms, and surgery is the standard treatment to remove the tumor, achieving a cure rate of over 70\%.
\end{itemize}

\begin{figure}[!ht]
    \centering
    \begin{tabular}{cccc}
         \includegraphics[width=0.22\textwidth]{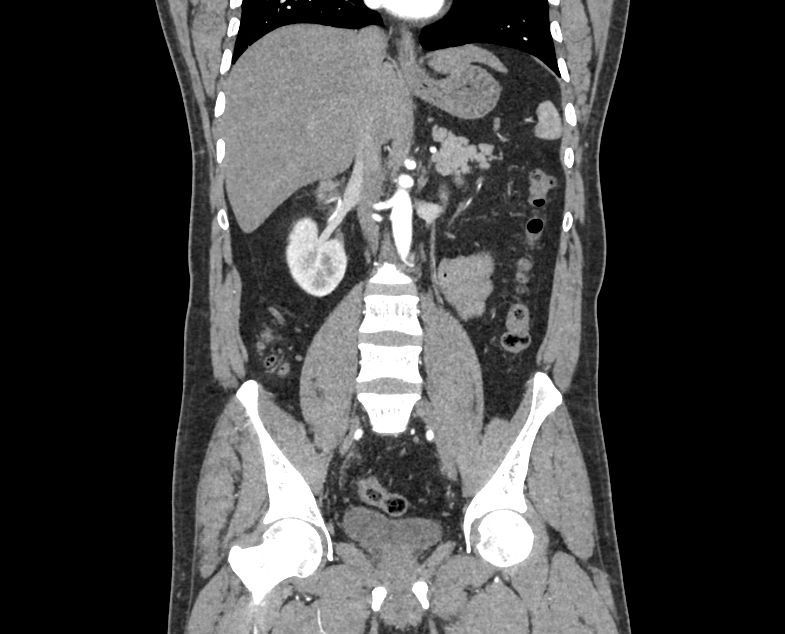} &  
         \includegraphics[width=0.22\textwidth]{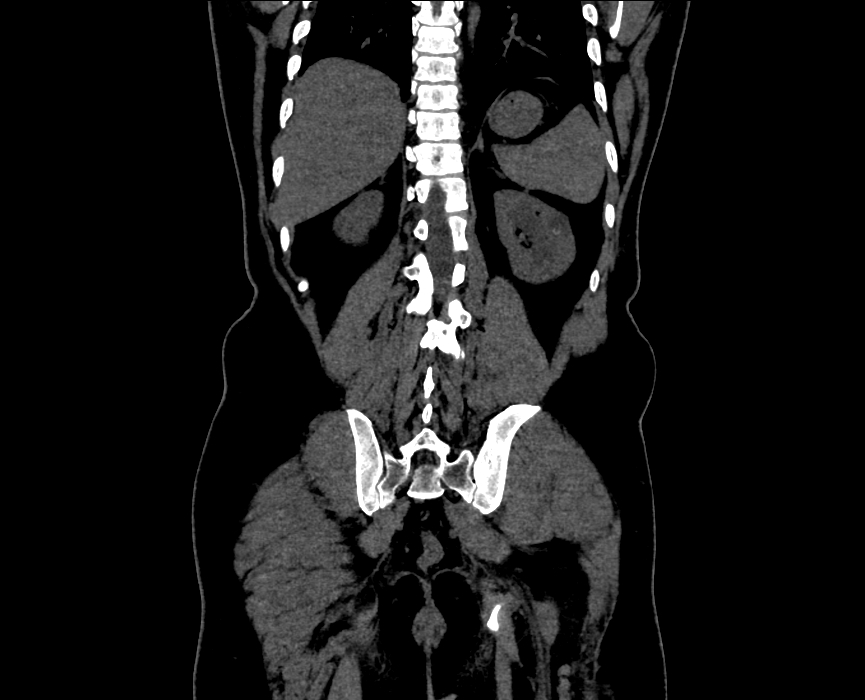}   &  
         \includegraphics[width=0.22\textwidth]{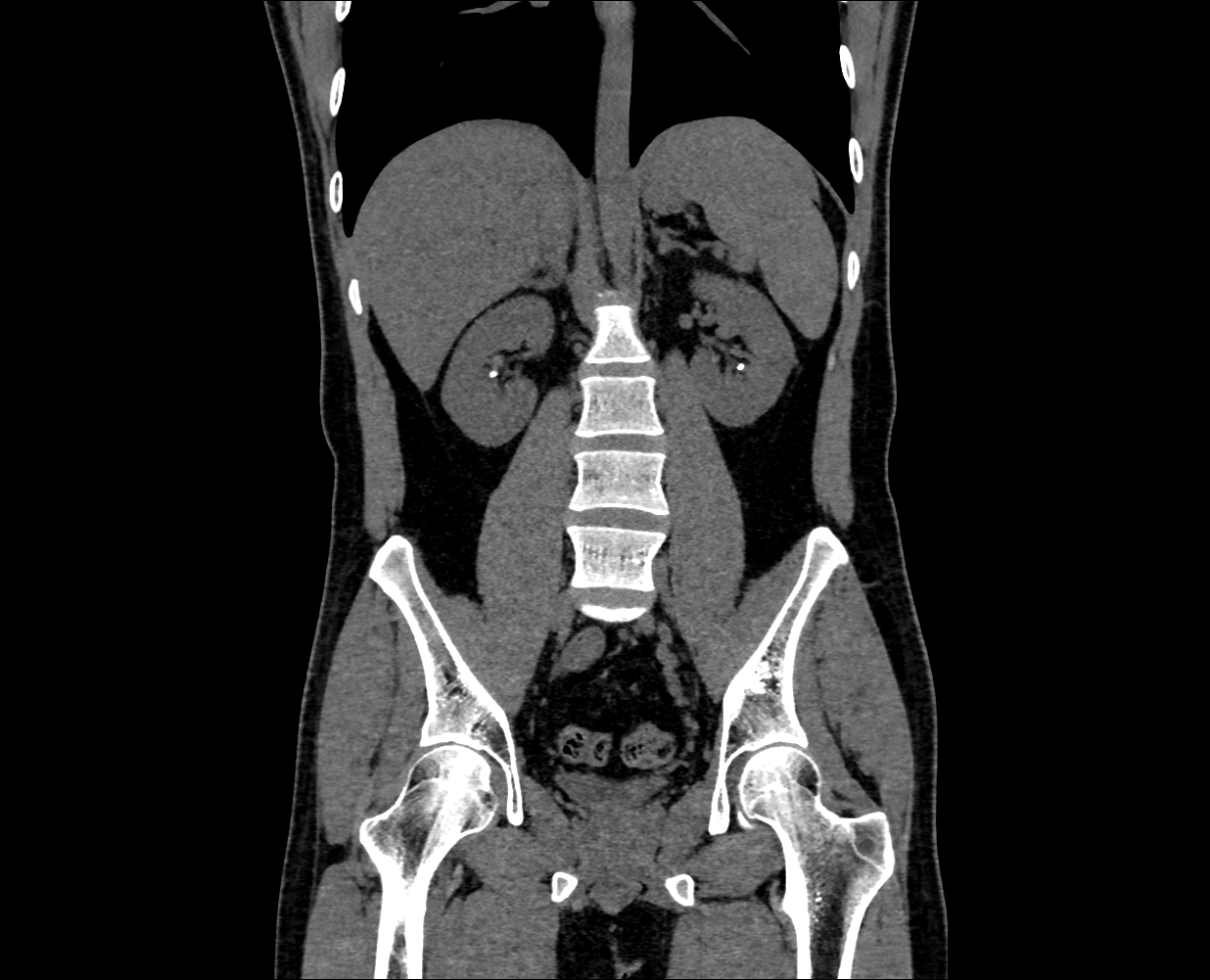}  &  
         \includegraphics[width=0.2\textwidth]{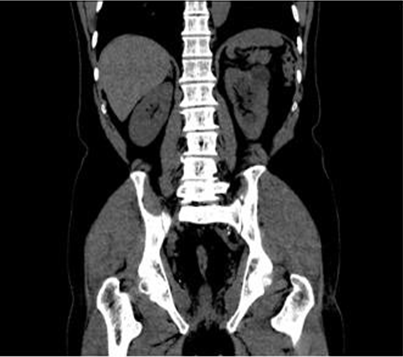}  \\ 
         \includegraphics[width=0.22\textwidth]{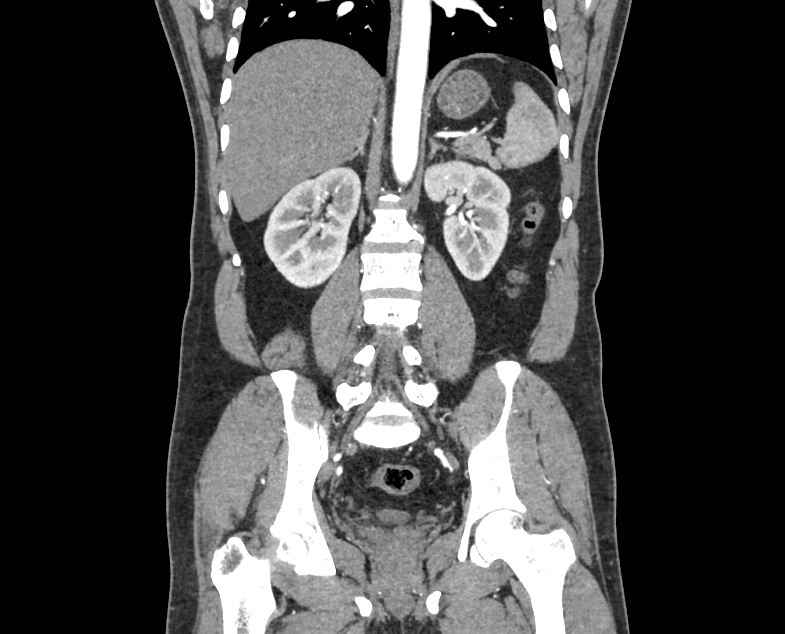} &
         \includegraphics[width=0.22\textwidth]{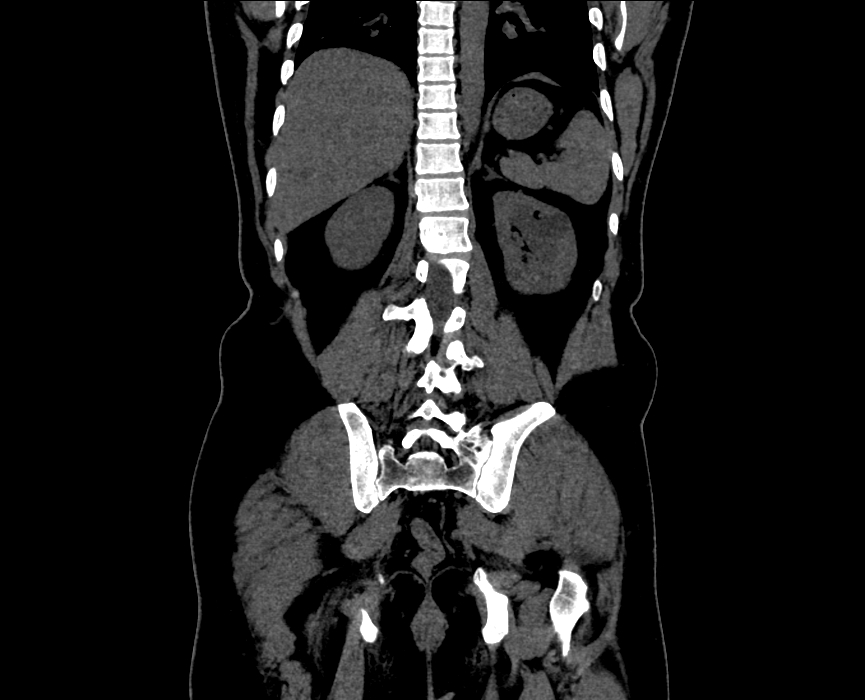}   &
         \includegraphics[width=0.22\textwidth]{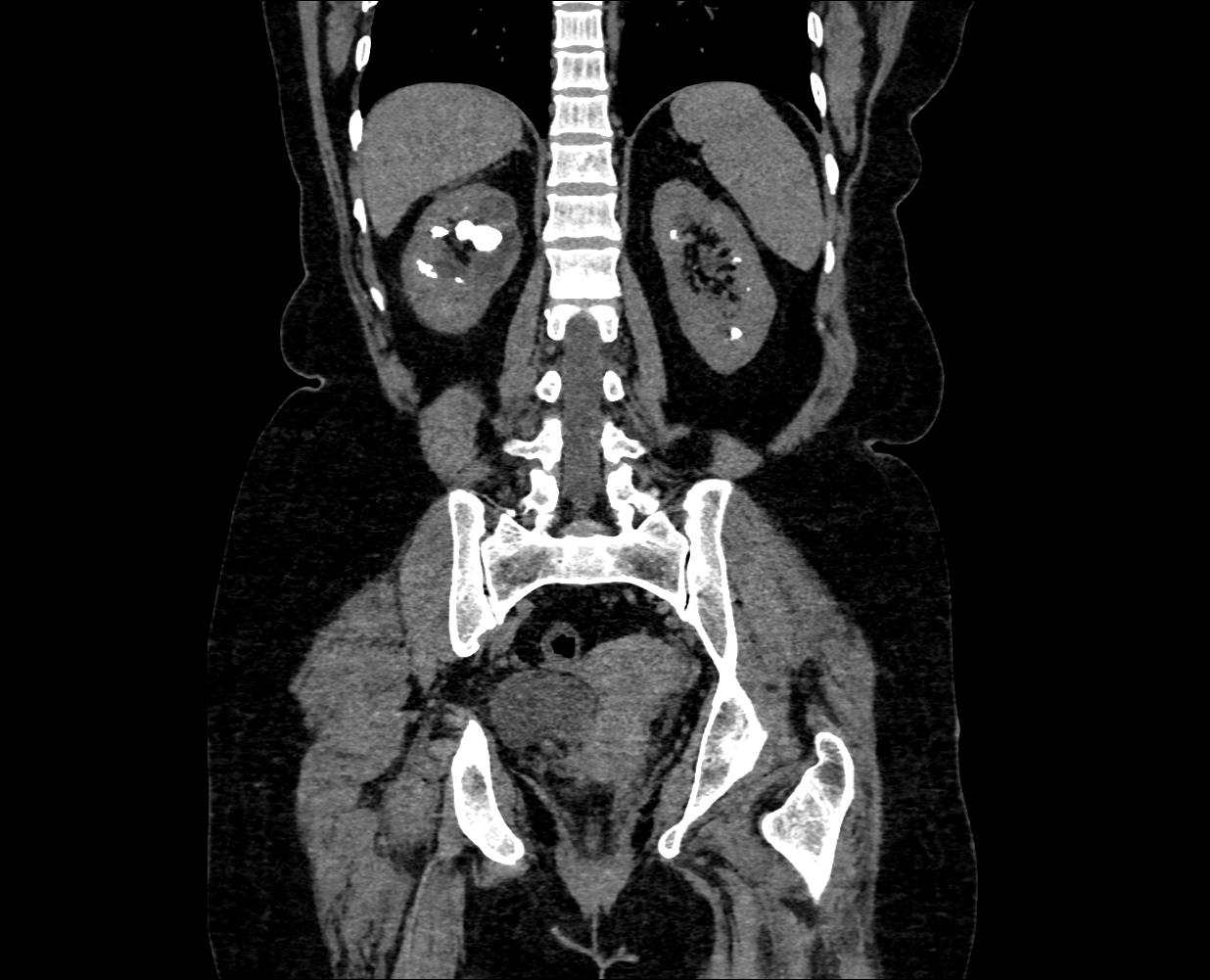}  &
         \includegraphics[width=0.2\textwidth]{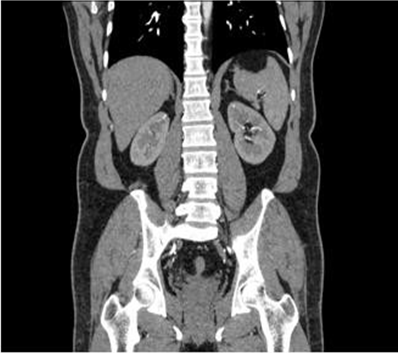}    
    \end{tabular}
    \caption{Examples from the kidney dataset. The two images in the first column represent CT images of \textit{normal} kidneys, the two images in the second column depict kidneys with \textit{cysts}, The images on the third column represent CT images of kidneys with \textit{stones}, and the images in the last column depict kidneys with \textit{cancer}.}
    \label{fi:kidney_tumors1}
\end{figure}

Figure \ref{fi:kidney_tumors1} shows several samples from the dataset, with two scans of each dataset class. 

Images of the dataset were selected from axial and coronal anatomical planes with and without contrast for the entire abdomen. The data format was DICOM and the images were converted to JPG format. The dataset originally contained 12,446 images, but we removed duplicate samples, reducing the total to 11,929 images. Table~\ref{tab:distribution_kidney_dataset} shows the distribution of images into the four classes.

\begin{table}[!ht]
    \caption{Distribution of samples in the classes of the kidney dataset. }
    \centering
    \begin{tabular}{lcc}
        \toprule
         Class & Number of images \\
         \midrule
         Normal & 5,002\\
         Cyst   & 3,284 \\
         Stone  & 1,360 \\
         Cancer & 2,283 \\
         \bottomrule
    \end{tabular}
    \label{tab:distribution_kidney_dataset}
\end{table}

Samples from the datasets were preprocessed by resizing all images to 224x224 pixels, converting DICOM files (for lung and kidney datasets) to JPEG format, and ensuring 3-channel RGB format for compatibility with Vision Transformers.

We used data augmentation techniques, such as rotation, flipping, and scaling, particularly for minority classes. However, the results did not show any improvement in this respect.

\subsection{Neural Networks}
In this work, we compare the performance of three vision Transformers, i.e., the Vision Transformer (ViT)~\cite{vit}, the Swin Transformer~\cite{swin}, and MaxViT~\cite{maxvit}, which have provided competitive results in many computer vision tasks.

\subsubsection{Vision Transformer}
The Vision Transformer (ViT) \cite{vit} is a model designed for image classification tasks, employing a Transformer-like architecture adapted for visual data. ViT begins by dividing an input image into fixed-size patches. This resembles how the original Transformer model breaks down text into tokens. Each image patch is flattened and transformed into a vector through linear embedding. This process converts the 2D patch into a 1D vector that the Transformer can process. Since the Transformer architecture does not inherently process sequential data, positional embeddings are added to the patch embeddings to retain the order of the patches, which is crucial for understanding the spatial relationships within the image. The sequence of patch embeddings, now with positional information, is fed into a standard Transformer encoder. The encoder consists of layers of multi-head self-attention and feed-forward neural networks.

For classification tasks, an extra learnable token, often referred to as the \textit{classification token} (CLS), is added to the sequence. The state of this token at the output of the Transformer encoder captures the global information about the image, which is used for the final classification.

The ViT architecture has shown remarkable abilities, achieving comparable or even better performance than traditional CNNs on many computer vision tasks. It represents a significant shift in how models process visual information, leveraging the power of self-attention mechanisms to capture global dependencies within the image.

\begin{table}[!ht]
    \centering
    \caption{Details of different variants of the ViT model.}
    \begin{tabular}{lccccc}
        \toprule
        Model &  Layers &   Hidden Layers & MLP size & Heads & Parameters (in Millions) \\
        \midrule
        ViT-Base & 12 &  768 & 3072 & 12 & 86 \\
        ViT-Large & 24 &  1024 & 4096 & 16 & 307 \\
        ViT-Huge & 32 & 1280 & 5120 & 16 & 632 \\
        \bottomrule
    \end{tabular}
    \label{tab:vit_models}
\end{table}

Table \ref{tab:vit_models} shows different configurations of the ViT model. Due to computational limitations, only the base model ViT-Base will be tested. Starting from this variant, we choose two alternatives, ViT-b-16 and ViT-b-32, for input patches of 16x16 pixels and 32x32 pixels, respectively. The model with smaller patches has a higher computational cost since the size of the Transformer sequence is inversely proportional to the square of the patch size.

\subsubsection{Swin Transformer}
The Swin Transformer~\cite{swin,swin2} is a type of Vision Transformer adapted for computer vision tasks, including image classification, object detection, and semantic segmentation. Unlike other vision Transformers that compute self-attention globally across the entire image, the Swin Transformer divides the image into smaller windows and applies self-attention within these local windows. This innovation significantly reduces computational complexity while maintaining its effectiveness. This architecture builds upon the ViT framework but introduces a hierarchical approach for processing images. It starts with smaller patches in the initial layers and progressively merges them into larger patches in deeper layers. This enables detailed image processing, capturing both local and global contexts.

It achieves linear computation complexity for input image size. This is in contrast to other architectures that have quadratic complexity due to global self-attention. The efficiency of the Swin Transformer makes it suitable as a general-purpose backbone for various vision tasks. Due to its hierarchical feature maps and efficient computation, the Swin Transformer is a versatile backbone for image classification tasks. It combines the power of Transformers with localized self-attention, making it an effective choice for processing medical images and detecting anomalies like tumors. Its benefits lie in improved efficiency, hierarchical feature extraction, and suitability for various computer vision tasks.

\subsubsection{MaxViT Transformer}

The MaxViT Transformer~\cite{maxvit} is a hybrid architecture that combines the strengths of CNNs and Vision Transformers to create a powerful image classification model. MaxViT integrates the inductive biases of CNNs with the global receptive field of ViTs. This combination allows it to achieve high performance across various parameters and metrics. The architecture introduces a multi-axis attention mechanism, incorporating blocked local and dilated global attention. This design enables the model to capture local and global spatial interactions at linear complexity, regardless of input resolution. Similar to traditional CNNs, MaxViT follows a hierarchical design. It builds upon a new type of basic building block that unifies MBConv~\cite{efficientnet} blocks and grid attention layers, allowing the model to reach a global context throughout the entire network. MaxViT scales well with large datasets and maintains linear complexity concerning the grid attention used, making it suitable for high-resolution inputs.

In the experiments, we will test different configurations of these architectures, like the ViT-b-16 and ViT-b-32 models. Regarding the Swin Transformer, there are three variants: \textit{tiny}, \textit{small}, and \textit{base}. In the case of MaxViT, we will test the base model. When considering performance, the focus is on the relationship between training time and final accuracy. Models that require extensive computation, both in terms of time and GPU memory load, were discarded.

\begin{table}[!ht]
    \centering
    \caption{Complexity of the models and its variants.}
    \begin{tabular}{lccc}
        \toprule
        Models & Parameters & FLOPS (in Millions) & Size (in Megabytes)\\
        \midrule
        ViT-b-16 & 86,415,592  & 456.96 & 360.24 \\
        ViT-b-32 & 88,185,064  & 456.96 & 344.59 \\
        Swin-t   & 28,288,354  & 123.43 & 141.22 \\
        Swin-s   & 49,606,258  & 208.40 & 236.33 \\
        Swin-b   & 87,768,224  & 363.69 & 397.41 \\
        MaxViT-b & 119,880,192 & 138.50  & 545.68 \\
        \bottomrule
    \end{tabular}
    \label{tab:models_complexity}
\end{table}

The variants of the models that we used in the experiments are presented in Table~\ref{tab:models_complexity}. For each model variant, we summarize the number of parameters, the floating point operations per second (FLOPS), and their size in megabytes.

\subsection{Experimental setup}
We use the Adam optimizer for learning the parameters of the neural networks, which is frequently used in image classification tasks.
For the loss function, we choose Categorical Cross-Entropy, given by the following expression:
\begin{equation}
    \mathcal{L}(\hat{y}_i,y_i)=-\sum\limits_{i=1}^N y_i\cdot \log \hat{y}_i, 
\end{equation}
where  $y_i$ is the true value of sample $i$, $\hat{y}_i$ is the prediction given by the neural network, and $N$ is the number of classes. This function yields a value between 0 and 1, representing a probability for each label or class trained in the model. When the error is very high, cross-entropy significantly penalizes those values that deviate from the expected predictions. 

The metrics used to compare the models are \textit{accuracy}, \textit{precision}, \textit{recall}, \textit{F1-score}, and AUC-ROC. These metrics rely on the true positive (TP), true negative (TN), false positive (FP), and false negative (FN) values of the classifications.

\textit{Accuracy} is an intuitive performance measure and is calculated as a ratio of correctly predicted observations to the total observations, and is given by:

\begin{equation}
accuracy = \frac{\text{TP}+\text{TN}}{\text{Total}}.
\end{equation}

\textit{Precision} measures the number of positive predictions that were correctly estimated and is calculated as:

\begin{equation}
precision=\frac{\text{TP}}{\text{TP}+\text{FP}}.  
\end{equation}

Recall or sensitivity measures the number of actual positives that were correctly identified. High recall means few false negatives and is calculated as:

\begin{equation}
recall=\frac{\text{TP}}{\text{TP}+\text{FN}}.
\end{equation}

F1-score is the harmonic mean between precision and recall. It balances the two metrics and is especially useful for imbalanced classes. It is calculated as:

\begin{equation}
F1-score=2\cdot\frac{precision \cdot recall}{precision+recall}.
\end{equation}

AUC-ROC is the Area Under the Receiver Operating Characteristic Curve and measures the area under the ROC curve, which is the relation between the True Positive Rate and the False Positive Rate. In the experiments, we also use confusion matrices to represent the misclassifications between the predicted and actual values of each class. 

The source code was implemented in Python and the deep learning models were implemented using the PyTorch framework. The training was carried out in an NVIDIA GPU GeForce RTX 2060 SUPER. The hyperparameters were tuned depending on the neural network architecture, dataset complexity, and available computational resources. We explored the following range of values:
\begin{itemize}
    \item Mini-batch Size: Ranges from 4 to 32 images per batch.
    \item Image Size: Images were scaled to a fixed dimension of 224x224 pixels.
    \item Image Channels: We used three channels (RGB).
    \item Learning Rate: The learning rate was set between $10^{-3}$ and $10^{-5}$, depending on the learning technique and the trained model.
    \item Dropout: In the final layer, neuron connections are deactivated at a rate of 30\% and 40\%. A dropout value within this range has always been active.
    \item Epochs: the number of epochs ranges from 5 to 200, depending on the dataset and mini-batch size applied.
\end{itemize}

To mitigate overfitting during training, we applied several strategies, like Dropout (30–40\%) on fully connected layers, early stopping based on validation loss, data augmentation (random rotations, flips, brightness adjustment), transfer learning with frozen layers, and reduced learning rates during fine-tuning phases. In our experiments, data augmentation did not contribute to improving the results, even for the less representative classes, and we decided not to include it in the final results.

\section{Results}
\label{se:results}

Each dataset was split into a training, validation, and test set. The training set was about 80\% of the dataset and was used to learn the parameters of the networks. The validation set was about 10\% of the dataset. This was used to validate the training process and to find appropriate hyperparameters. The test set was about 10\%. This was used to evaluate the performance of models. 

\subsection{Results with transfer learning}

In this section, we analyze the global performance of the models for each dataset and in combination. We trained the models using transfer learning in two steps: in the first one, we only optimized the parameters of the head of the models; in the second step, we optimized all the parameters of the networks with a small learning rate.

\begin{table}[!ht]
    \centering
    \caption{Accuracy of the models after the first step of transfer learning. This table shows the accuracy of the models for each dataset. The last row shows the results of each model by combining the three datasets. Bold letters highlight the best results in each row.}
    \begin{tabular}{lcccccc}
        \toprule
        Dataset & ViT-b-32 & Swin-t & MaxViT-b  \\
        \midrule
            Brain	      &  \textbf{93.27\%} & 93.07\% & 91.47\% \\
            Kidney	      &  \textbf{97.15\%} & 95.55\% & 88.51\% \\
            Lung	      &  \textbf{93.33\%} & 90.48\% & 87.62\% \\
            All datasets  &  \textbf{95.43\%} & 93.82\% & 83.60\% \\
        \bottomrule
    \end{tabular}
    \label{tab:training_scratch}
\end{table}

Table~\ref{tab:training_scratch} shows the results at the end of the first step. This table compares the accuracy of the three neural networks across the individual and combined datasets. The results indicate that ViT-b-32 consistently outperforms the other two models across all scenarios. It achieves the highest accuracy on the Brain (93.27\%), Kidney (97.15\%), and Lung (93.33\%) datasets individually. Moreover, its performance remains strong even when trained on the combined dataset, achieving 95.43\% accuracy. This suggests that ViT-b-32 is particularly robust and generalizes well across different types of tumor data.

The Swin-t model performs similar to ViT-b-32 in most cases. On the Brain dataset, its accuracy (93.07\%) is nearly equal to that of ViT-b-32. However, there is a slightly larger performance gap in the kidney (95.55\%) and lung (90.48\%) datasets. When evaluated on the combined dataset, Swin-t achieves an accuracy of 93.82\%, which, although lower than ViT-b-32, still indicates relatively strong generalization capabilities. 

In contrast, the MaxViT-t model shows significantly lower performance across all datasets. It presents particularly weak results on the kidney tumor dataset (88.51\%). Its performance drops when trained on the combined dataset, where it achieves only 83.60\% accuracy. These results suggest that MaxViT-t may struggle to capture relevant features effectively in this domain or may require more task-specific fine-tuning to achieve competitive results.

The results of the second step show a notable improvement across all models and datasets when compared to the initial training results, indicating that further optimization significantly enhances performance; see Table ~\ref{tab:transfer_learning}. This comparison provides key insights into how each model benefits from additional fine-tuning.

\begin{table}[!ht]
    \centering
    \caption{Accuracy of the models using transfer learning. This table shows the accuracy of the ViT-b-32, Swin-t, and MaxViT-b for each dataset. The last row shows the results of each model by combining the three datasets. Bold letters highlight the best results in each row.}
    \begin{tabular}{lccc}
        \toprule
        Dataset & ViT-b-32 & Swin-t & MaxViT-b \\
        \midrule
        Brain	       & 97.07\% & \textbf{99.53\%} & 99.27\% \\
        Kidney	       & 97.73\% & \textbf{99.75\%} & 99.75\% \\
        Lung	       & 95.24\% & \textbf{97.14\%} & 94.29\% \\
        All datasets   & 97.03\% & \textbf{99.43\%} & 98.68\% \\
        \bottomrule
    \end{tabular}
    \label{tab:transfer_learning}
\end{table}

On the Brain dataset, the accuracy of ViT-b-32 rises from 93.27\% to 97.07\%, and on the Kidney dataset from 97.15\% to 97.73\%. On the Lung dataset, it improves from 93.33\% to 95.24\%, and on the combined dataset, the accuracy moves up from 95.43\% to 97.03\%. Its relative ranking drops as the other two models show even greater gains.

The Swin-t model shows the most significant improvements. On the Brain dataset, its accuracy increases from 93.07\% to 99.53\%, and on Kidney tumors, from 95.55\% to 99.75\%. For Lung tumors, its accuracy goes from 90.48\% to 97.14\%, and on the combined dataset, from 93.82\% to 99.43\%. These substantial gains position Swin-t as the top performer across all categories after fine-tuning. The model overtakes ViT-b-32 with significant margins, demonstrating high adaptability to the tumor datasets.

MaxViT-t shows a remarkable recovery after the second fine-tuning step. Its Brain tumor classification accuracy jumps from 91.47\% to 99.27\%, and its Kidney dataset performance leaps from 88.51\% to 99.75\%, matching Swin-t. Lung tumor classification improves from 87.62\% to 94.29\%, and the combined dataset result increases significantly from 83.60\% to 98.68\%. This transformation indicates that MaxViT-t may require more extensive training to unlock its full potential, possibly due to its complex architecture or deeper capacity.

The performance of the three neural networks on the Kidney dataset reveals important distinctions not only in terms of accuracy, but also when considering model size, parameter count, and computational cost (FLOPS), as depicted in Table~\ref{kidney_precision}. 

\begin{table}[!ht]
  \caption{Performance assessment of different configurations of vision Transformers for the kidney dataset: The first column shows the version of the model used in each experiment; the second column depicts the accuracy of the model for the test set; the third column shows the number of parameters of each network; the fourth column, the number of FLOPS; and the last column, the size of the networks in megabytes. The last two columns are copied from Table~\ref{tab:models_complexity}.}
  \centering
  \begin{tabular}{lcccc}
    \toprule
    Model & Accuracy & Number of Parameters & FLOPS (in Millions) & Size (in Megabytes) \\
    \midrule
        ViT-b-32  &  97.73\% & 88,185,064 & 456.96 & 344.59 \\
        Swin-t    &  99.75\% & 28,288,354 & 123.43 & 141.22 \\
        MaxViT-t  &  99.75\% & 30,919,624 & 5600.0 & 118.80 \\
    \bottomrule
  \end{tabular}
  \label{kidney_precision}
\end{table}

Both Swin-t and MaxViT-t achieve the highest accuracy of 99.75\%, outperforming ViT-b-32. Swin-t stands out for its excellent balance between performance and efficiency. It delivers top-tier accuracy with fewer parameters (28.3 million), a relatively low FLOP count of 123.43M, and a moderate model size of 141.22 MB. This indicates that Swin-t is highly optimized for lightweight and fast inference, making it ideal for applications with limited computational resources, such as edge devices or clinical settings where rapid predictions are essential.

MaxViT-t matches Swin-t in accuracy but with substantially higher computational demand. Its FLOPS soar to 5600M, nearly 45× greater than Swin-t, despite having a similar number of parameters (30.9M) and even a smaller model size (118.80 MB). This suggests that MaxViT-t leverages more complex internal operations, likely due to its hybrid use of convolutional and transformer blocks. Its high FLOP makes it less efficient for real-time deployment unless high-performance hardware is available.

ViT-b-32 has the largest number of parameters (88.2M), the largest model size (344.59 MB), and is less accurate. Additionally, its FLOPS are between Swin-t and MaxViT-t, which suggests diminishing returns in terms of both accuracy and efficiency. Therefore, we may conclude that Swin-t offers the best trade-off between accuracy and efficiency, MaxViT-t is highly accurate but computationally expensive, and ViT-b-32, while accurate, is less efficient in both size and performance.

\subsection{Results for individual datasets}

Given the results of the previous section, we provide more details about the performance of each model and dataset using all the metrics.  

\begin{table}[!ht]
  \caption{Performance of the models for the Brain dataset. Bold letters indicate the best result in each metric.}
  \centering
  \begin{tabular}{lccccc}
    \toprule
    Model & Accuracy (\%) & Precision & Recall & F1-score & AUC-ROC \\
    \midrule
        ViT-b-32 & 97.07 & 0.971 & 0.971 & 0.971 & 0.998 \\
        Swin-t	 & \textbf{99.53} & \textbf{0.995} & \textbf{0.996} & \textbf{0.995} & \textbf{1.0} \\
        MaxViT-t & 99.27 & 0.993 & 0.993 & 0.993 & \textbf{1.0} \\
    \bottomrule
  \end{tabular}
  \label{tab:brain_results}
\end{table}

The results for the Brain tumor dataset reveal excellent performance across all three models, with Swin-t and MaxViT-t outperforming ViT-b-32, especially in terms of classification precision and general robustness; see Table~\ref{tab:brain_results}.

\begin{figure}[!ht]
    \centering
    \begin{subfigure}[t]{0.32\textwidth}
        \centering
        \includegraphics{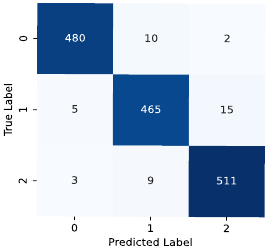}
        \caption{ViT-b-32}
    \end{subfigure}%
    ~ 
    \begin{subfigure}[t]{0.32\textwidth}
        \centering
        \includegraphics{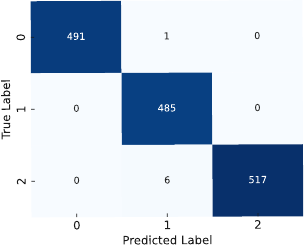}
        \caption{Swin-t}
    \end{subfigure}
    ~ 
    \begin{subfigure}[t]{0.32\textwidth}
        \centering
        \includegraphics{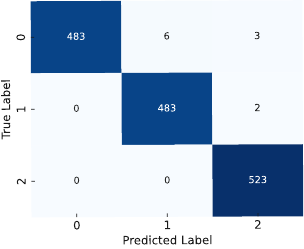}
        \caption{MaxViT-t}
    \end{subfigure}
    \caption{Confusion matrices for the models using the Brain dataset. The label codes are Glioma (0), Meningioma (1), and Pituitary (2).}
    \label{fi:cf_brain}
\end{figure}

The precision, recall, and F1-score (0.971) of ViT-b-32 are aligned, suggesting that the model is useful for detecting brain tumors without favoring one type of error. Swin-t stands out as the top performer on this dataset, with a high AUC-ROC, indicating flawless discrimination between tumor classes. Its precision (0.995) and recall (0.996) are both exceptionally high, leading to an F1-score of 0.995, which reflects high reliability and minimal error rate. MaxViT-t also performs exceptionally well, with precision and recall producing a high F1-score of 0.993, and an AUC-ROC showing excellent class separability. Although marginally behind Swin-t in terms of overall metrics, MaxViT-t is still highly effective for this classification task. These results reflect a few misclassifications in the confusion matrices of Figure\ref{fi:cf_brain}.

The performance results for the Kidney dataset in Table~\ref{tab:kideny_results} reflect that ViT-b-32 also stands behind the other two models. Its precision (0.983) and recall (0.964) indicate that it misses more true cases. The AUC-ROC of 0.997 is high, showing the model’s good overall discrimination ability. Swin-t is the top performer with a precision of 0.997 and a recall of 0.996. This performance, especially given its relatively lightweight architecture, emphasizes Swin-t’s remarkable efficiency and robustness for kidney tumor classification.

\begin{figure}[!ht]
    \centering
    \begin{subfigure}[t]{0.32\textwidth}
        \centering
        \includegraphics{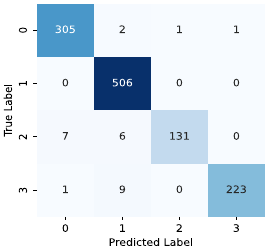}
        \caption{ViT-b-32}
    \end{subfigure}%
    ~ 
    \begin{subfigure}[t]{0.32\textwidth}
        \centering
        \includegraphics{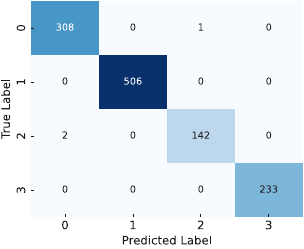}
        \caption{Swin-t}
    \end{subfigure}
    ~ 
    \begin{subfigure}[t]{0.32\textwidth}
        \centering
        \includegraphics{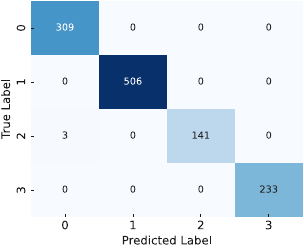}
        \caption{MaxViT-t}
    \end{subfigure}
    \caption{Confusion matrices for the models using the Kidney dataset. The label codes are Glioma (0), Meningioma (1), and Pituitary (2).}
    \label{fi:cf_kidney}
\end{figure}

MaxViT-t matches Swin-t in accuracy and slightly surpasses it in precision (0.998), indicating even fewer false positives. However, its recall (0.995) is marginally lower than Swin-t’s, though the difference is negligible in practice. Its AUC-ROC confirms that it attains a highly effective class separation. Their confusion matrices in Figure~\ref{fi:cf_kidney} reflect the low misclassification rates of these two models. 

\begin{table}[!ht]
  \caption{Performance of the models for the Kidney dataset. Bold letters indicate the best result in each metric.}
  \centering
  \begin{tabular}{lccccc}
    \toprule
    Model & Accuracy (\%) & Precision & Recall & F1-score & AUC-ROC \\
    \midrule
        ViT-b-32 & 97.73 & 0.983 & 0.964 & 0.972 & 0.997 \\
        Swin-t   & \textbf{99.75} & 0.997 & \textbf{0.996} & \textbf{0.996} & \textbf{1.0} \\
        MaxViT-t & \textbf{99.75} & \textbf{0.998} & 0.995 & \textbf{0.996} & \textbf{1.0} \\
   \bottomrule
  \end{tabular}
  \label{tab:kideny_results}
\end{table}

Table \ref{tab:lung_results} shows the results for the Lung dataset, which has far fewer samples than the other datasets and presents imbalanced classes. This experiment also demonstrates the outstanding performance of Swin-t in terms of overall accuracy and balance across evaluation metrics. ViT-b-32 recall (0.879) is noticeably lower, suggesting the model misses more true cases. Swin-t has the highest accuracy of 97.14\%, with good precision (0.974) and recall (0.917). These values indicate that the Swin-t predictions are correct and also capture more actual tumor cases. Its AUC-ROC of 0.993 further reinforces its excellent class separation ability. 

MaxViT-t, on the other hand, shows the lowest performance on this dataset, with an accuracy of 94.29\% and the lowest scores in all other metrics. Its precision (0.898) and recall (0.877) indicate more false positives and negatives compared to the other models. The F1-score of 0.883 suggests a weaker balance in classification, and the AUC-ROC of 0.979 ranks below both ViT-b-32 and Swin-t. 

\begin{figure}[!ht]
    \centering
    \begin{subfigure}[t]{0.32\textwidth}
        \centering
        \includegraphics{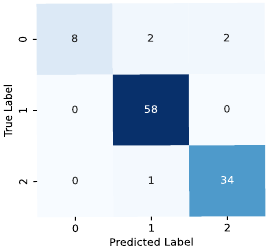}
        \caption{ViT-b-32}
    \end{subfigure}%
    ~ 
    \begin{subfigure}[t]{0.32\textwidth}
        \centering
        \includegraphics{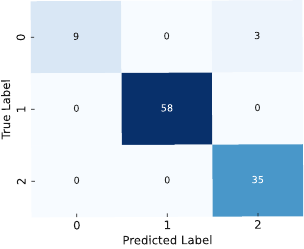}
        \caption{Swin-t}
    \end{subfigure}
    ~ 
    \begin{subfigure}[t]{0.32\textwidth}
        \centering
        \includegraphics{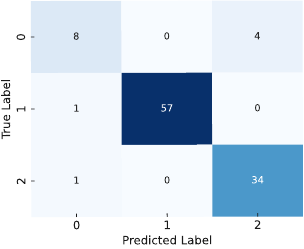}
        \caption{MaxViT-t}
    \end{subfigure}
    \caption{Confusion matrices for the models using the Lung dataset. The label codes are Glioma (0), Meningioma (1), and Pituitary (2).}
    \label{fi:cf_lung}
\end{figure}

The confusion matrices in Figure~\ref{fi:cf_lung} show that Swin-t has fewer misclassifications, and only in the third class.

\begin{table}[!ht]
  \caption{Performance of the models for the Lung dataset. Bold letters indicate the best result in each metric.}
  \centering
  \begin{tabular}{lccccc}
    \toprule
    Model & Accuracy (\%) & Precision & Recall & F1-score & AUC-ROC \\
    \midrule
        ViT-b-32 & 95.24 & 0.965 & 0.879 & 0.911 & 0.982 \\
        Swin-t & 97.14 & 0.974 & 0.917 & 0.939 & 0.993 \\
        MaxViT-t & 94.29 & 0.898 & 0.874 & 0.883 & 0.979 \\
   \bottomrule
  \end{tabular}
  \label{tab:lung_results}
\end{table}

We may conclude that Swin-t is the best overall performer when we use separate datasets, as it combines high accuracy, robust generalization, and balanced metrics across all three tumor datasets. It also adapts better to smaller datasets with imbalanced classes.

\subsection{Results for the combined dataset}

When we combine all the datasets and train the models, we also obtain high performance in general. Looking at Table~\ref{tab:lung_results}, the results show that Swin-t provides the best overall performance across all metrics.

\begin{figure}[!ht]
    \centering
    \includegraphics[width=0.8\textwidth]{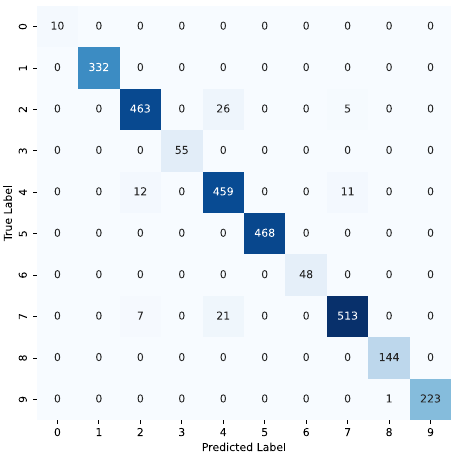}
    \caption{Confusion matrix for the ViT-32-b model using the combined datasets. The label codes are: Benign (0 –  Lung), Cyst (1 – Kidney), Glioma (2 – Brain), Malignant (3 – Lung), Meningioma (4 – Brain), Normal (5 – Kidney), Normal (6 – Lung), Pituitary (7 – Brain), Stone (8 – Kidney), Tumor (9 – Kidney).}
    \label{fi:cf_combined_vit}
\end{figure}

ViT-b-32 provides competitive results, especially in precision and recall, and a high AUC-ROC metric. It presents slightly more misclassifications in several classes (e.g., Glioma and Meningioma), possibly due to class similarity or overlap; see Figure ~\ref{fi:cf_combined_vit}.

\begin{figure}[!ht]
    \centering
    \includegraphics[width=0.8\textwidth]{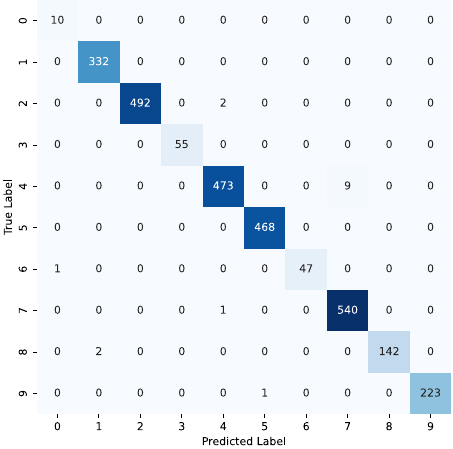}
    \caption{Confusion matrix for the Swin-t model using the combined datasets. The label codes are: Benign (0 –  Lung), Cyst (1 – Kidney), Glioma (2 – Brain), Malignant (3 – Lung), Meningioma (4 – Brain), Normal (5 – Kidney), Normal (6 – Lung), Pituitary (7 – Brain), Stone (8 – Kidney), Tumor (9 – Kidney).}
    \label{fi:cf_combined_swin}
\end{figure}

The confusion matrix of the Swin-t model, in Figure~\ref{fi:cf_combined_swin}, reveals a few off-diagonal misclassifications. It has the highest recall and F1-score, indicating a strong ability to capture true positives while maintaining precision.
 
\begin{table}[!ht]
  \caption{Performance of the models when we combine all the datasets. Bold letters indicate the best result in each metric.}
  \centering
  \begin{tabular}{lccccc}
    \toprule
    Model & Accuracy (\%) & Precision & Recall & F1-score & AUC-ROC \\
    \midrule	
    ViT-b-32 & 97.03 & 0.983 & 0.983 & 0.983 & 0.999 \\
    Swin-t   & \textbf{99.43} & \textbf{0.988} & \textbf{0.994} & \textbf{0.991} & \textbf{1.0} \\
    MaxViT-t & 98.68 & 0.974 & 0.984 & 0.978 & \textbf{1.0} \\
   \bottomrule
  \end{tabular}
  \label{tab:combined_results}
\end{table}

\begin{figure}[!ht]
    \centering
    \includegraphics[width=0.8\textwidth]{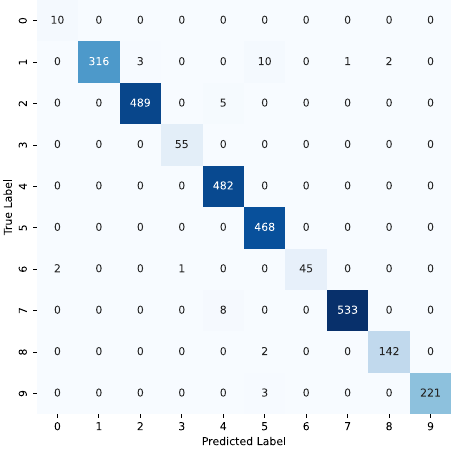}
    \caption{Confusion matrix for the MaxViT-t model using the combined datasets. The label codes are: Benign (0 –  Lung), Cyst (1 – Kidney), Glioma (2 – Brain), Malignant (3 – Lung), Meningioma (4 – Brain), Normal (5 – Kidney), Normal (6 – Lung), Pituitary (7 – Brain), Stone (8 – Kidney), Tumor (9 – Kidney).}
    \label{fi:cf_combined_maxvit}
\end{figure}

MaxViT-T also provides good performance, although behind the Swin-t models in all metrics and ViT-b-32 in overall scores. Its confusion matrix (Figure~\ref{fi:cf_combined_maxvit}) shows misclassification, particularly between similar or adjacent classes. It has a high AUC-ROC, which suggests good separability in probability estimates.

It is interesting to note that the results for the less representative classes, with few samples, are satisfactory, obtaining a high rate of correct classifications. This behavior is similar for the three architectures. This highlights the capacity of Vision Transformers to deal with imbalanced data. Besides, we observe that the classes of the Lung dataset have fewer misclassifications if we compare with its independent counterpart, which indicates that it takes advantage of the contents of the other datasets.

\begin{table}[!ht]
  \caption{Comparison of the performance of the models when trained with individual and combined datasets. The results of the individual datasets from the previous tables have been averaged (avg). Bold letters indicate the best result in each metric.}
  \centering
  \begin{tabular}{lccccc}
    \toprule
    Model & Accuracy (\%) & Precision & Recall & F1-score & AUC-ROC \\
    \midrule
    ViT-b-32 Individual (avg) & 96.77 & 0,975 & 0.949 & 0.959 & 0.994 \\
    ViT-b-32 Combined & 97.03 & 0.983 & 0.983 & 0.983 & 0.999 \\
    Swin-t Individual (avg) & 98.96 & \textbf{0.988} & 0.975 & 0.980 & 0.998 \\
    Swin-t Combined & \textbf{99.43} & \textbf{0.988} & \textbf{0.994} & \textbf{0.991} & \textbf{1.0} \\
    MaxViT-t Individual (avg) & 97.99 & 0.966 & 0.961 & 0.962 & 0.995 \\
    MaxViT-t Combined & 98.68 & 0.974 & 0.984 & 0.978 & \textbf{1.0} \\
   \bottomrule
  \end{tabular}
  \label{tab:individual_combined_results}
\end{table}

Finally, Table \ref{tab:individual_combined_results} compares the results when the models are trained with independent datasets and when trained with the combined dataset. We averaged the results of the independent datasets. In this case, ViT-b-32 shows a significant improvement when trained on the combined dataset, suggesting it benefits from more diverse training data.

The Swin-t model shows the best performance in both scenarios, but slightly better when trained on the combined dataset. MaxViT-T already performs strongly with individual datasets, but it still has a small gain from training on the combined data.

Training on the combined dataset generally yields better or at least equivalent performance for all three models, especially for ViT-b-32, which appears more sensitive to training data diversity.

\section{Conclusion}
\label{se:conclusion}

This work presented a comprehensive evaluation of three state-of-the-art Vision Transformer architectures --—ViT-b-32, Swin-t, and MaxViT-t--— applied to the classification of Brain, Lung, and Kidney tumors using MRI and CT scans. Through extensive experimentation using individual and combined datasets, we analyzed each model’s performance in terms of accuracy, precision, recall, F1-score, and AUC-ROC.

The results of the first fine-tuning step indicated that ViT-b-32 initially offered the most consistent performance across all datasets, particularly benefiting from diverse training data. However, after the second and more extensive fine-tuning phase, Swin-t emerged as the most effective model, outperforming the others in every metric and dataset configuration. It achieved the highest accuracy (99.75\%) and maintained a low computational footprint, making it suitable for deployment in clinical environments with limited resources.

MaxViT-t, while initially underperforming, demonstrated significant gains after full fine-tuning, especially in the Kidney and Brain datasets. However, its high computational cost makes it less practical for real-time applications, despite its high accuracy. ViT-b-32, although competitive in early stages and particularly strong on recall and generalization, was outperformed by the other models in accuracy and efficiency.

Training with a combined dataset led to improved generalization and robustness, particularly for ViT-b-32 and MaxViT-t, which benefited from the added data diversity. This was even more important for less-represented and imbalanced classes. This highlights the potential of Vision Transformers to handle multimodal and multiclass medical imaging tasks within a unified framework. 

This study confirms that Transformer-based architectures are highly effective for medical image classification, especially when fine-tuned appropriately. Swin-t stands out as the best trade-off between performance and efficiency, making it a strong candidate for future deployment in computer-aided diagnosis systems. 

In future work, we will explore the integration of other imaging modalities, attention interpretability, and real-time inference optimization to further improve the models’ clinical applicability.

\bibliographystyle{unsrt}  
\bibliography{references}  

\end{document}